\documentclass{article}

\usepackage{microtype}
\usepackage{graphicx}
\usepackage{subfigure}
\usepackage{booktabs} 

\usepackage{hyperref}

\usepackage[accepted]{arxiv}

\usepackage{amsmath}
\usepackage{amssymb}
\usepackage{mathtools}
\usepackage{amsthm}
\usepackage{multicol}
\usepackage{multirow}
\usepackage{enumitem}

\usepackage[capitalize,noabbrev]{cleveref}

\theoremstyle{plain}

\theoremstyle{definition}

\theoremstyle{remark}

\newcommand{\update}[1]{\textcolor{black}{#1}}

\usepackage[textsize=tiny]{todonotes}
\usepackage{lipsum}

\icmltitlerunning{Take the Bull by the Horns: Hard Sample-Reweighted Continual Training Improves LLM Generalization}

\begin{document}

\twocolumn[
\icmltitle{Take the Bull by the Horns: \\Hard Sample-Reweighted Continual Training Improves LLM Generalization}



\begin{icmlauthorlist}
\icmlauthor{Xuxi Chen}{yyy}
\icmlauthor{Zhendong Wang}{yyy}
\icmlauthor{Daouda Sow}{xxx}
\icmlauthor{Junjie Yang}{xxx}
\icmlauthor{Tianlong Chen}{zzz}
\icmlauthor{Yingbin Liang}{xxx}
\icmlauthor{Mingyuan Zhou}{yyy}
\icmlauthor{Zhangyang Wang}{yyy}

\end{icmlauthorlist}

\icmlaffiliation{yyy}{University of Texas at Austin}
\icmlaffiliation{xxx}{The Ohio State University}
\icmlaffiliation{zzz}{University of North Carolina at Chapel Hill}

\icmlcorrespondingauthor{Xuxi Chen}{xxchen@utexas.edu}

\icmlkeywords{Machine Learning, ICML}

\vskip 0.3in
]



\printAffiliationsAndNotice{\icmlEqualContribution} 

\newcommand{\fix}{\marginpar{FIX}}
\newcommand{\new}{\marginpar{NEW}}
\newcommand{\xx}[1]{{\color{blue} XX: #1}}
\newcommand{\ds}[1]{{\color{green} DS: #1}}

\begin{abstract}

In the rapidly advancing arena of large language models (LLMs), a key challenge is to enhance their capabilities amid a looming shortage of high-quality training data. Our study starts from an \textit{empirical strategy} for the light continual training of LLMs using their original pre-training data sets, with a specific focus on selective retention of samples that incur \textbf{moderately high} losses. These samples are deemed informative and beneficial for model refinement, contrasting with the highest-loss samples, which would be discarded due to their correlation with data noise and complexity. We then formalize this strategy into a \textit{principled framework} of Instance-Reweighted Distributionally Robust Optimization (\textbf{IR-DRO}). IR-DRO is designed to dynamically prioritize the training focus on informative samples through an instance reweighting mechanism, streamlined by a closed-form solution for straightforward integration into established training protocols. Through rigorous experimentation with various models and datasets, our findings indicate that our sample-targeted methods significantly improve LLM performance across multiple benchmarks, in both continual pre-training and instruction tuning scenarios. Our codes are available at \url{https://github.com/VITA-Group/HardFocusTraining}.

\end{abstract}
\vspace{-1em}
\section{Introduction}

Large Language Models (LLMs) have demonstrated an impressive ability to understand and reason in multiple tasks and have shown some surprising abilities such as in-context learning~\citep{brown2020language,wei2022chain}. These models require training with an extensive corpus of data, including web-crawled documents~\citep{raffel2020exploring} and scholarly articles~\citep{gao2020pile}. 
For example, the pretraining of a \texttt{LLaMA} model consumes up to one trillion word tokens, and \texttt{GPT-4}~\citep{achiam2023gpt} reportedly uses more than $13$ trillion tokens for pretraining. 


However, these increasing numbers of tokens deployed for pretraining LLMs have become a concern. 
It has been suggested that the depletion of high-quality data sources may become an increasingly pressing challenge, with projections indicating a potential shortfall in the supply of such data resources by 2026~\citep{villalobos2022will}. While high-quality synthetic data can help mitigate the shortage of data in some downstream tasks~\citep{he2022synthetic,mitra2023orca}, it has been recognized that they often have domain shifts~\citep{sankaranarayanan2018learning} and biases~\citep{belgodere2023auditing}, making the high-quality of ``real'' training data still preferred for LLMs training. Recent studies~\citep{gunasekar2023textbooks,li2023textbooks,javaheripi2023phi} have also emphasized the critical role of carefully selected, high-quality data in enhancing LLM performance.
These observations therefore necessitate a research question: \textit{Given an already trained LLM, how can we further improve its performance through light continual training, by strategically exploiting the samples from the same training dataset?} Addressing this can conserve computational efforts already invested in LLMs and increase performance without additional data. 


To answer the question aforementioned regarding the continual training of LLMs, we begin our investigation by examining the samples associated with high and small loss values. Our results, detailed in Section~\ref{sec:example}, reveal a noteworthy observation: \textit{the traditional strategy that always selects samples with the highest losses can impair LLM performance after continuous training}. d Manual examination of examples reveals that these samples are more likely to be associated with noise or inherently challenging patterns, such as those found in online advertisements and reference lists from papers.
To address this issue, we propose an empirical strategy for selecting samples that exhibit a moderately high level of loss. Our proof-of-concept experiments demonstrate that applying this strategy for continual pretraining enhances performance on various evaluation benchmark datasets.


Despite the effectiveness of this empirical strategy, it requires trial and error to choose proper hyperparameters in a model-specific way. 
In response, we present a more principled algorithm, called \textbf{I}nstance-\textbf{R}eweighted \textbf{D}istributionally \textbf{R}obust \textbf{O}ptimization (IR-DRO), to automatically identify and prioritize informative samples that are beneficial to the continual training without the tedious hyperparameter tuning process. Additionally, we have developed a closed-form formulation that enables the straightforward integration of IR-DRO into the existing training pipelines of LLMs.

Our experiments demonstrate that IR-DRO outperforms our empirical strategy at various benchmarks. Furthermore, we illustrate that our proposed techniques are versatile and can improve other training scenarios, including instruction tuning.
We summarize our contribution as below:\vspace{-1em} 
\begin{itemize}[leftmargin=*]
    \item To identify samples beneficial for the continual training of LLMs, we introduce an empirical algorithm designed to select instances with moderate complexity levels. This approach is based on our observations on textual patterns and has proven to be effective in enhancing the performance of continually trained models.

    \item We further \update{adapt and develop} IR-DRO, a principled optimization-based algorithm that utilizes the instance reweighting formulation to dynamically identify and prioritize crucial and informative samples for the continual training of LLMs.
 
    \item Extensive experiments conducted on various pre-trained models, such as \texttt{OPT}~\citep{zhang2022opt}, \texttt{Sheared-LLaMA}~\citep{xia2023sheared}, and \texttt{LLaMA}~\citep{touvron2023llama}, have confirmed the effectiveness of our proposed algorithms. Our methods achieve improved performance on evaluation benchmarks after continual pre-training and instruction tuning, respectively.\vspace{-0.5em} 
\end{itemize}

\vspace{-1em}
\section{Related Works}
\vspace{-0.5em}
\subsection{Pretraining of LLMs}
\vspace{-0.5em}
Current state-of-the-art LLMs are usually pretrained on billions or even trillions of tokens~\citep{touvron2023llama}, supported by the increasing size of the underlying pretraining datasets. \texttt{BERT}~\citep{devlin2018bert} was pretrained on \texttt{BookCorpus}~\citep{zhu2015aligning} and English Wikipedia, which have 3 trillion words in total. \texttt{GPT-2}~\citep{radford2019language} was pre-trained on a data set crawled on the web named \texttt{WebText} that had about $40$ GB of text data. \texttt{T5}~\citep{raffel2020exploring} was pretrained on Colossal Clean Crawled Corpus (\texttt{C4}), a cleaned subset of the \texttt{Common Crawl}'s web crawl corpus. More recently, \texttt{LLaMA}~\citep{touvron2023llama} had enlarged its pretraining datasets to encompass multiple domains such as \texttt{Common Crawl}, \texttt{C4}, and source codes from Github, resulting in over one trillion of tokens. The increasing size of the pretraining corpus makes it infeasible to ensure the quality of collected samples, which poses new data quality challenges to the (pre-)training of LLMs~\citep{naveed2023comprehensive}. \citet{trinh2018simple} has identified that there existed a substantial portion of documents ``whose contents are mostly unintelligible'' in the \texttt{Common Crawl}. All these challenges motivate us to find a better strategy to utilize massive yet under-curated data. 

\subsection{Data Re-weighting and Selection}


It has been known that i.i.d sampling may not be the best strategy when training with large and noisy data potentially coming from compositional domains. 
DoReMi~\citep{xie2023doremi} proposes using an auxiliary model to determine the optimal weights for different domain data and achieve better performance. Skill-It~\citep{chen2023skill} introduces an online data sampling technique that dynamically updates the mixture of data corpus from predefined skill sets. 
Some seek methods with a finer granularity, i.e., reweighting data at the instance level. These works generally use the loss values~\citep{loshchilov2015online,jiang2019accelerating} or the gradient norms~\citep{katharopoulos2018not} to derive a probability distribution for data sampling. However, these methods often require a second forward-backward process that damages the scalability of the methods. \citet{fan2023irreducible} uses an auxiliary model and defines a learnability score for sample reweighting, yet the auxiliary model needs to undergo a similar pretraining process, which incurs additional training costs. \update{The assignment of weights to individual samples can also be conceptualized within the context of optimization problems. A special family of these problems, the KL-regularized distributionally robust optimization (DRO) in a min-max form, can be efficiently addressed by transforming into simpler compositional minimization forms~\citep{qi2021online,qi2022stochastic}. Follow-up works have demonstrated that such a formulation can mitigate issues related to data imbalance~\citep{qi2023attentionalbiased} and enhance contrastive learning by customizing temperatures for individual samples~\citep{qiu2023not}. }

Dataset selection or pruning has also been an attractive research direction as it identifies data samples that are more important to the training process. It has been used in efficient training~\citep{mirzasoleiman2020coresets,killamsetty2021grad,killamsetty2021glister} and active learning~\citep{fu2013active,sener2017active}. 
For example, \citet{attendu2023nlu} suggests pruning datasets based on predicted probabilities and actual labels of downstream tasks. Similarly, \citet{marion2023less} explores the pruning of the data sets for pre-training LLMs. In our study, our aim is to identify the samples that contribute the most effectively to the continuing ongoing training of LLMs.

\vspace{-1.0em}
\section{MidRanking: An Empirical Strategy of Loss Ranking-based Sample Selection}
\vspace{-0.5em}
\label{sec:heuristic}

\subsection{Preliminaries} 
\vspace{-0.5em}
Let us denote the input sample (\textit{i.e.}, a sentence) by $\boldsymbol{x}=[t_1,t_2,\dots t_n]\in\mathbb{N}^{n}$ where $t_i$ indicates the $i$-th token, the model parameters by $\boldsymbol{\theta}$, and the loss objective by $\mathcal{L}$. In this paper, we adopt the training objective of auto-regressive language models, which has the form of
$
\mathcal{L}(\boldsymbol{\theta};\boldsymbol{x})=\sum_{i=2}^{n}-\log \mathbb{P}(t_i | t_{<i}),
$ where $\mathbb{P}(t_i | t_{<i})$ represents the predicted probability of the $i$-th token being $t_i$, given the first $i-1$ tokens as the input. Fine-tuning LLMs on downstream tasks often relies on the empirical risk minimization (ERM), which is mathematically expressed as:
\begin{equation}
\label{eqn:basic}
\small{
\min_{\boldsymbol{\theta}} L(\boldsymbol{\theta}) = \frac{1}{|\mathcal{B}|} \sum_{i=1}^{|\mathcal{B}|} \mathcal{L}(\boldsymbol{\theta}; \boldsymbol{x}_i),}
\end{equation}
where $|\mathcal{B}|$ represents the batch size and $\boldsymbol{x}_i$ represents the $i$-th instance in the batch $\mathcal{B}$.

In this work, our primary focus is on the \textit{continual training} scenario, where a pretrained LLM undergoes additional training using data instances from the original training dataset
This setting is practical and economical, as it allows us to reuse the community's efforts in pretraining LLMs and further enhance the models' performance without requiring samples from additional data sources.

\subsection{Critical Sample Selection:  Is the `Largest Loss' Necessarily the 'Most Informative'?} 
\label{sec:example}
The speed at which neural networks process and learn from data is not uniform, which has led to research into the concept of curriculum learning~\citep{bengio2009curriculum}. Some data samples are quickly understood and retained by the network early in training. In contrast, the network may only remember more complex samples in the later stages of training. This suggests that these harder-to-learn samples might not be fully learned if training is concluded prematurely, especially considering the prevailing view that most modern LLMs are not sufficiently trained~\citep{radford2019language}. Therefore, persistently focusing on these hard samples could enable models to better absorb those overlooked information, potentially improving performance.



Intuitively, higher loss values in a model's output often reflect greater confusion, signaling that the model struggles to grasp the patterns or information in the more challenging samples. Yet, it's critical to recognize that high loss values may arise from several sources. \underline{First}, these samples could be riddled with noise or contain patterns that are inherently difficult to predict, which presents a challenge for effective learning. Take, for instance, web-sourced ads that tend to repeat phrases, or bibliographies from academic articles that contain elements of randomness. \underline{Second}, these samples might encompass essential information that the model has not yet fully learned. This includes examples rich in complex mathematical concepts or advanced medical jargon. Prioritizing ongoing training with such challenging data can potentially yield marked performance enhancements.

Our objective is to establish a method that can discern between two distinct origins of data. We begin by conducting an observational study to assess the loss values of samples within the \texttt{C4}  dataset using the pre-trained \texttt{LLaMA-v2-7B} model. The results, as shown in Table~\ref{tab:loss}, include selected samples and their respective loss values, leading to two principal insights: (1) the sample that registers the highest loss values tends to be mostly "noisy," as exemplified by its intricate reference list structure; (2) a sample with moderate loss appears to carry more semantic value, offering a comprehensive explanation of psychology and its related fields. This evidence indicates that the largest loss samples are more prone to represent noise or inherent ambiguity rather than informational content, compared to samples with moderate losses. Consequently, training exclusively on these highest-loss samples could potentially hinder model performance.

\begin{table}[ht]
    \centering
    \vspace{-1em}
    \caption{Data instances in \texttt{C4} associated with high and medium levels of loss by \texttt{LLaMA-2-7B}. The selected sample with the highest loss corresponds to a segment of the reference in a paper, while the sample with medium loss appears more informative. }
    \begin{tabular}{p{0.75\linewidth}|p{0.1\linewidth}}
    \toprule
        Sampled Text & Loss (Rank)\\
        \midrule
         ``Esbjörn-Hargens, S. (2010). An Integral Overview of Climate Change: Why Truth is not Enough. Journal of Integral Theory and Practice, 5(1), 1-42.
Ballard, D., Reason, P., \& Coleman, G. (2010). Using the AQAL Framework to Accelerate Responses to Climate Change. Journal of Integral Theory and Practice, $\dots$'' [truncated] & $1.1293$ $(1)$\\
         \midrule
         ``The term psychology is generally suggested to describe behavioral procedures that relate to the feelings or the mind. The term emotional reliance is generally indicated to define the emotional and also mental procedures that are related to the development of, and also $\dots$'' [truncated] & $0.5350$ $(5000)$ \\ 
         \bottomrule
    \end{tabular}
    \label{tab:loss}
\end{table}

To address this challenge and validate our hypothesis, we introduce an ad-hoc algorithm named \textit{MidRanking}. It is designed to improve the continual training of LLMs through adaptive sample selection.  Rather than exclusively focusing on those with the highest losses, MidRanking employs a tactical method, ranks samples by their loss values in descending order. The algorithm then bypasses the top $n_1|\mathcal{B}|$ samples, presumed to be the noisiest, and begins to train on the following
$n_2|\mathcal{B}|$ samples. This approach, applied batch-wise, streamlines the training process. Removes the burdensome requirement of recalculating loss values for the entire dataset after each update, a procedure that would otherwise require substantial computational resources.

\begin{table*}[ht]
    \centering
    \caption{Performance on multiple benchmarks using \texttt{OPT-125M} and \texttt{OPT-350M} under the continual pretraining setting. The introduction of evaluation benchmarks are in Section~\ref{sec:settings}. }
    \label{tab:sanity}
    \begin{tabular}{c|c|cccccc|c}
    \toprule \multirow{2}{*}{Model} & \multirow{2}{*}{Method} & \multicolumn{6}{c|}{Evaluation Benchmarks} & \multirow{2}{*}{Average}\\
    
     & & ARC-C & HellaSwag & PiQA & WinoGrande & BoolQ & MMLU \\
    \midrule
        \multirow{5}{*}{\texttt{OPT-125M}} & Original & 22.78 & 31.33 & 62.02 & 50.28 & 55.41 & 26.21 & 41.34 \\
        & Uniform & 22.87 & 31.35 & 62.40 & 49.88 & 55.84 & 26.12 & 41.41\\ 
        \cmidrule{2-9} & HighRanking & 22.78 & 31.30 & 62.30 & 50.20 & 55.84 & 26.16 & \textbf{41.43} \\
        & MidRanking & 22.87 & 31.36 & 62.30 & 50.12 & 55.81 & 26.11 & \textbf{41.43} \\
        & LowRanking & 22.70 & 31.41 & 62.08 & 49.64 & 56.02 & 26.08 & 41.32 \\
        \midrule
        \multirow{5}{*}{\texttt{OPT-350M}} & Original & 23.98 & 36.77 & 64.69 & 52.33 & 57.65 & 26.06 & 43.58 \\
        & Uniform & 24.49 & 36.81 & 64.74 & 52.80 & 58.32 & 26.05 & 43.87\\ 
        \cmidrule{2-9} & HighRanking & 24.49 & 36.67 & 64.64 & 52.96 & 58.40 & 26.01 & 43.86 \\
        & MidRanking & 24.83 & 36.74 & 64.80 & 52.57 & 58.60 & 26.37 & \textbf{43.99}  \\
        & LowRanking & 24.15 & 36.66 & 64.80 & 52.33 & 58.20  & 26.19 & 43.72\\
        \bottomrule
    \end{tabular}
    \vspace{-0.5em}
\end{table*}


\subsection{Experimental Validation of MidRanking}
\label{exp:naive}
To assess the efficacy of the MidRanking approach, we conduct a series of proof-of-concept experiments on \texttt{OPT-125M} and \texttt{OPT-350M} models. We use \texttt{C4} as the dataset to sample training data. A total of $100$ batches are sampled from the data set used to continuously pre-train these models. In our experiments, we set the batch size to be $8$, and set $n_1$ and $n_2$ to be $0.25$. 
We also compare MidRanking against various baselines using different settings of $n_1$ and $n_2$ to simulate several typical scenarios: (1) \textit{HighRanking}, where $n_1$ is set to $0$ and $n_2$ remains $0.25$. This is the same as selecting only the samples with largest losses; (2) \textit{LowRanking}, where $n_1$ is set to $0.75$ and $n_2$ remains $0.25$, which is identical to selecting only the samples with smallest losses; and (3) \textit{Uniform}, where $n_1$ is $0$ and $n_2$ is $1$, mirroring the standard fine-tuning algorithm. Introduction of evaluation benchmarks is deferred to Section~\ref{sec:settings}. 

Table~\ref{tab:sanity} showcases MidRanking's consistent effectiveness.
Multiple variant analysis shows that LLM performance can be improved through continual training in samples from the same pretraining dataset, with MidRanking achieving the most notable improvements: it consistently exceeds all baselines with the \texttt{OPT-125M} model and shows a more significant boost with the \texttt{OPT-350M} model, where it exceeds the baseline models by up to $0.41$ points in average scores. Further experiments with different architectures, such as \texttt{Sheared-LLaMA} as detailed in Table~\ref{tab:performance_on_opt_ir_dro}.




\vspace{-0.2em}
\section{IR-DRO: Principled Optimization-based Selection with an Efficient Solution} 
\vspace{-0.2em}

Although MidRanking successfully enhances the performance of pretrained models, its reliance on empirical observations introduces certain limitations. Specifically, the algorithm's effectiveness depends on two manually selected hyperparameters, $n_1$ and $n_2$. Optimizing these parameters through a grid search can be computationally intensive, especially for large models. Moreover, the optimal settings for these parameters are not universally applicable and must be determined individually. Dropping samples solely based on losses can also be too coarse for continual training. These drawbacks underscore the necessity for a more systematic approach to identifying critical samples.





\begin{figure*}[ht]
    \centering
    \includegraphics[width=0.9\linewidth]{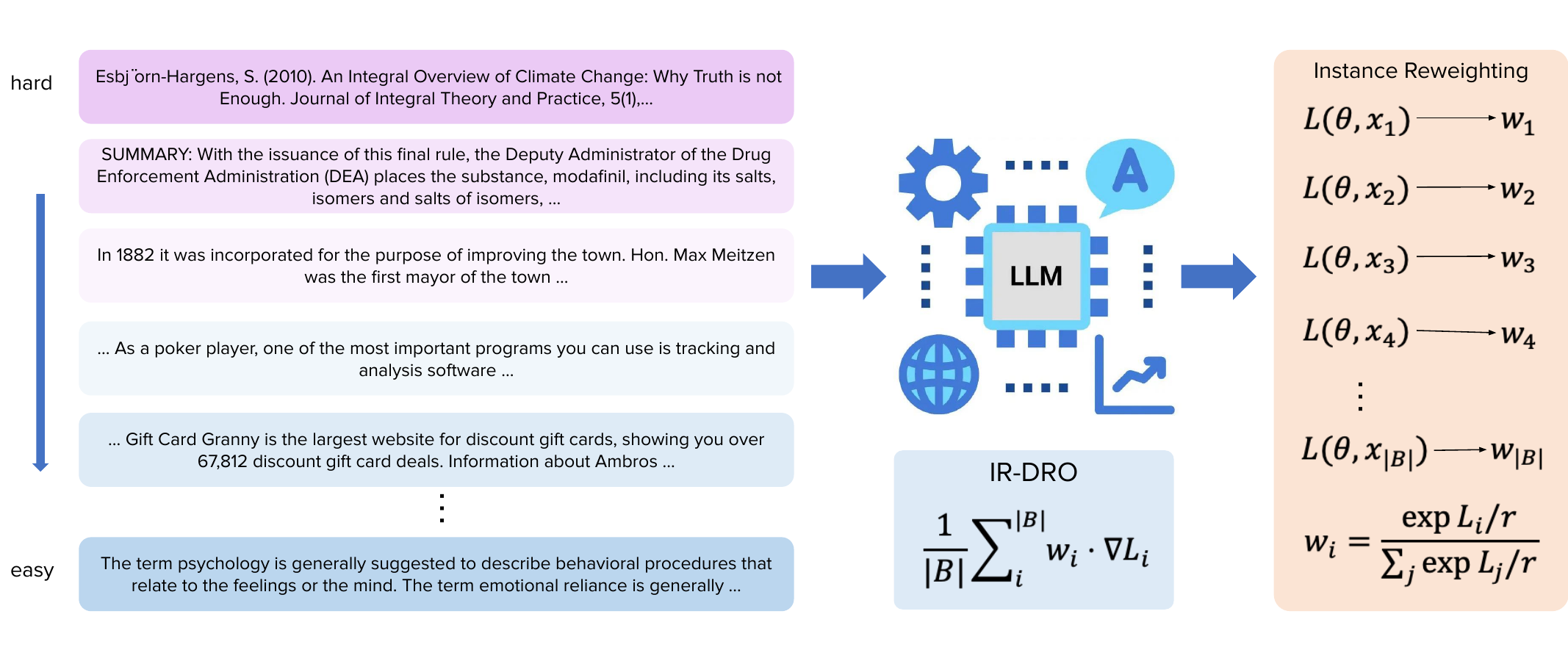}
    \vspace{-1em}
    \caption{We introduce IR-DRO, a principled optimization-based framework named that automatically decides the importance scores at instance level and reweight the training process.}
    \label{fig:framework}
    \vspace{-1em}
\end{figure*}

\vspace{-0.5em}
\subsection{Overview} 
\vspace{-0.5em}
We consider an instance-level reweighting scheme for solving this problem. Essentially, we extend Equation~\ref{eqn:basic} by introducing a weight coefficient for each sample in a batch. The optimization problem has become the following form: 
\begin{align*}
    &\min_{\theta} \frac{1}{|\mathcal{B}|}\sum_{i=1}^{|\mathcal{B}|} w_i \mathcal{L}(\boldsymbol{\theta}; \boldsymbol{x}_i) \quad \text{s.t.} \quad \sum_{i=1}^{|\mathcal{B}|} w_i = 1, w_i \geq 0,
\end{align*} where $w_i$ represents the assigned weight to the $i$-th data instance in the batch. This formulation facilitates methods that prioritize certain data by strategically designing instance weights, thereby bypassing the cumbersome hyperparameter selection inherent in MidRanking. Although it appears straightforward, determining $w$ is often challenging for LLMs, and current re-weighting strategies for LLMs usually depend on proxy or meta-models.
These methods, despite their conceptual appeal, present notable disadvantages: (1) \textbf{Increased training complexity}: The addition of proxy or meta-models to LLMs significantly elevates the overall training complexity, making these approaches less feasible for continual training settings, particularly with resource-intensive LLMs; (2) \textbf{Lack of theoretical grounding}: Many existing re-weighting methods lack a rigorous theoretical foundation for computing the coefficients of instances, making it difficult to analyze their behavior and generalization guarantees. 

\vspace{-0.5em}
\subsection{Formulation} 
\vspace{-0.5em}
We \update{adapt an optimization framework, named \textbf{I}nstance-\textbf{R}eweighted \textbf{D}istributionally \textbf{R}obust \textbf{O}ptimization (IR-DRO), based on the formulation of distributionally robust optimization~\citep{qi2021online,qi2022stochastic}}. It overcomes the limitations of existing methods while offering several key advantages: \begin{itemize}[leftmargin=*]
\vspace{-0.5em}
    \item \textbf{On-the-fly instance-level re-weighting}: Unlike techniques that rely on proxy models, our method applies re-weighting directly during the training of LLMs, eliminating the need for additional models and thus significantly reducing the complexity of training.. 
    \item \textbf{Principled instance weights via DRO}: We utilize the robustness of Distributionally Robust Optimization (DRO) to determine optimal instance weights in a systematic way, ensuring that training prioritizes challenging examples.
    \item \textbf{Closed-form instance weights}: We have developed a closed-form solution for the instance weight optimization problem within the DRO framework, facilitating efficient computation of instance weights and seamless integration during training.
\end{itemize} 

Specifically, we formulate a minimax optimization problem with a KL-divergence regularizer \update{inspired by \citet{qi2021online}} as follows: 
\begin{equation} \label{eq:dro}
\min_{\boldsymbol{\theta}} \max_{\boldsymbol{w} \in \mathcal{P}} \sum_{i=1}^N w_i \mathcal{L}(\boldsymbol{\theta}; \boldsymbol{x}_i) - r \sum_{i=1}^N w_i \log(Nw_i)
\end{equation}
where $\mathcal{P}$ represents the probability simplex, ensuring valid weight distributions, and $r$ controls the strength of the KL-divergence regularization. This formulation incentivizes the LLM to focus on hard cases while minimizing loss across all data points.
Incorporating a KL regularizer is crucial for balancing uniform and worst-case weight distributions, thus preventing overfitting to single ``worst" examples.  Without it, the objective function would linearly depend on the weight vector $w$, resulting in a simple one-hot vector solution (characterized by a solitary ``1'' at the index $i$ corresponding to the highest loss) for the ``max'' problem in Equation~\ref{eq:dro}. 
Such a one-hot vector approach, however, can destabilize the optimization process and negatively affect the model's ability to generalize across the training distribution, as noted in~\citet{qian2019robust,wang2021adversarial}. Furthermore, as observed in Section~\ref{exp:naive}, excessively prioritizing samples with the largest losses can degrade performance.



One of the key advantages of employing the KL-divergence as the regularizer is the emergence of a unique, closed-form solution to the maximization problem in Equation~\ref{eq:dro} (see Section~\ref{app:proof} for proof). Specifically, this yields the unique optimal weights vector $\boldsymbol{w}^*$ with its $i$-th entry given by 
\begin{align} \label{eq:wstar}
    w^*_i = \frac{\exp\left(\frac{\boldsymbol {\mathcal{L}}_i(\boldsymbol{\theta})}{r}\right)}{\sum_j \exp\left(\frac{\boldsymbol {\mathcal{L}}_j(\boldsymbol{\theta})}{r}\right)}, 
\end{align}
where $\boldsymbol {\mathcal{L}}_i(\boldsymbol{\theta}) = \mathcal{L}(\boldsymbol{\theta}; \boldsymbol{x}_i)$ indicates the loss objective calculated on the $i$-th data point. By substituting $\boldsymbol{w}^*$ back into the original objective in Equation~\ref{eq:dro}, the problem \update{can be transformed into the following compositional optimization problem~\citep{qi2021online}}: 
\begin{align} \label{eq:scp}
    &\underset{\boldsymbol{\theta}}\min ~ r\log \left(\frac{1}{N} \sum_{i=1}^N \exp\left(\frac{\boldsymbol {\mathcal{L}}_i(\boldsymbol{\theta})}{r}\right) \right). 
\end{align}
Stochastic algorithms have been devised to solve the above compositional problems. In particular, \citet{wang2017stochastic} proposed the stochastic compositional gradient descent (SCGD) method, which stands out as the first stochastic algorithm to address these problems, characterized by its straightforward implementation. Notably, SCGD simply keeps a running average of the inner-level function in the composition objective to address the biases of the stochastic sample gradient due to the non-linear sampling probabilities in Equation~\ref{eq:dro}. Concurrently, the inclusion of this regularizer promotes smooth coefficients and \textbf{prevents the reliance solely on samples with the highest losses}. These strategies align with our aforementioned empirical observations.

\paragraph{Practical algorithm.} We have developed a direct and practical algorithm to address the optimization challenge. Specifically, our approach uses the optimal weights derived in Equation~\ref{eq:wstar} to solve the instance reweighted problem as presented in Equation~\ref{eq:dro}, bypassing the need for reformulation in Equation~\ref{eq:scp}. It is important to note that the optimal weights in Equation~\ref{eq:wstar} are dependent on the model parameters $\boldsymbol{\theta}$. Nevertheless, our findings indicate that the algorithm's performance remains consistent regardless of whether this dependency is considered. Consequently, we omit this dependency in all subsequent experiments. The details of our method are documented in Algorithm~\ref{alg:practical}.

\begin{algorithm}

    \caption{Practical Algorithm for IR-DRO}
    \label{alg:practical}
    \begin{algorithmic}
        \STATE {\bfseries Input:} stepsize $\alpha$, initialization $\boldsymbol \theta_0$. 
		\FOR{$k=0,1,...,T-1$}
		\STATE{Draw a minibatch of samples $\mathcal{B} = \{\boldsymbol x_i\}_i$}
        \FOR{each $\boldsymbol x_i \in \mathcal{B}$ (in parallel)} 
        \STATE{Compute sample loss $\boldsymbol {\mathcal{L}}_i(\boldsymbol{\theta}_t)$ and gradient $\nabla \boldsymbol {\mathcal{L}}_i(\boldsymbol{\theta}_t)$}
        \STATE{Obtain sample weight $w_i^*$ using \cref{eq:wstar}} 
		\ENDFOR
        \STATE{Compute gradient $\nabla \boldsymbol {\mathcal{L}}(\boldsymbol{\theta}_t) = \frac{1}{|\mathcal{B}|} \sum_{i=1}^{|\mathcal{B}|} w_i^* \nabla \boldsymbol {\mathcal{L}}_i(\boldsymbol{\theta}_t)$} 
        \STATE{Update $\boldsymbol \theta_{t+1} = \boldsymbol \theta_{t} - \alpha \nabla \mathcal{L}(\boldsymbol{\theta}_t)$}
        \ENDFOR
    \end{algorithmic}
\end{algorithm}

\begin{table*}[ht]
    \centering
    \caption{Performance on multiple benchmarks with \texttt{OPT-125M}, \texttt{OPT-350M}, \texttt{Sheared-LLaMA-1.3B}, and \texttt{LLaMA-v2-7B}. }
    \label{tab:performance_on_opt_ir_dro}
    \resizebox{0.95\linewidth}{!}{
    \begin{tabular}{c|c|cccccc|c}
    \toprule \multirow{2}{*}{Model} & \multirow{2}{*}{Method} & \multicolumn{6}{c|}{Evaluation Benchmarks} & \multirow{2}{*}{Average}\\
    
     & & ARC-C & HellaSwag & PiQA & WinoGrande & BoolQ & MMLU \\
    \midrule
        \multirow{4}{*}{\texttt{OPT-125M}} & Original & 22.78 & 31.33 & 62.02 & 50.28 & 55.41 & 26.21 & 41.34 \\
        & Uniform & 22.87 & 31.35 & 62.40 & 49.88 & 55.84 & 26.12 & 41.41  \\ 
        \cmidrule{2-9} & MidRanking & 22.87 & 31.36 & 62.30 & 50.12 & 55.81 & 26.11 & 41.43\\
        & IR-DRO & 22.87 & 31.37 & 62.35 & 50.36 & 55.80 & 26.12 & \textbf{41.48} \\
        \midrule
        \multirow{4}{*}{\texttt{OPT-350M}} & Original & 23.98 & 36.47 & 64.69 & 52.33 & 57.65 & 26.06 & 43.58 \\
        & Uniform & 24.49 & 36.81 & 64.74 & 52.80 & 58.32 & 26.05 & 43.87\\ 
        \cmidrule{2-9} & MidRanking & 24.83 & 36.74 & 64.80 & 52.57 & 58.60 & 26.37 & 43.99 \\
        & IR-DRO & 24.56 & 36.94 & 64.74 & 52.96 & 58.72 & 26.31 &\textbf{44.04} \\
        \midrule
        \multirow{4}{*}{\texttt{Sheared-LLaMA-1.3B}} & Original & 29.10 & 59.35 & 73.39 & 58.09 & 61.96 & 25.32 & 51.20  \\
        & Uniform & 30.46 & 60.26 & 74.10 & 59.19 & 60.76 & 25.28 & 51.68 \\ 
        \cmidrule{2-9} & MidRanking & 30.29 & 60.28 & 73.98 & 59.67 & 60.98 & 25.18 & 51.73\\
        & IR-DRO & 30.38 & 60.34 & 74.02 & 59.57 & 61.10 & 25.31 & \textbf{51.79} \\
         \midrule \multirow{3}{*}{\texttt{LLaMA-v2-7B}} & Original & 46.25 & 76.03 & 79.11 & 69.14 & 77.68 & 45.50 & 65.62 \\
        & Uniform & 46.01 & 76.04 & 79.02 & 68.94 & 77.71 & 45.38 & 65.52\\ 
        & IR-DRO & 46.13 & 76.06 & 79.17 & 69.08 & 77.92 & 45.92 & \textbf{65.72}\\
        \bottomrule
    \end{tabular}}
    
\end{table*}
\vspace{-0.5em}
\section{Experiments}
\vspace{-0.2em}
\subsection{Experimental Settings}
\vspace{-0.2em}

\label{sec:settings}

\textbf{Baselines.} We mainly compare IR-DRO against three baseline methods: (1) \textit{Original}, where we directly evaluate the pretrained models; (2) \textit{Uniform}, where the samples in every continual batch are assigned equal weights. This corresponds to the simplest fine-tuning baseline; (3) \textit{MidRanking}, as we introduced in Section~\ref{sec:heuristic}, where we manually select samples with medium-level losses for training.

\textbf{Tasks.} We conduct experiments in the following scenarios: (1) \textit{continual pretraining}, whose setting is the same as the one we presented in Section~\ref{exp:naive}; and (2) \textit{continual instruction-tuning}, which involves initially performing standard supervised fine-tuning using a specific instruction tuning dataset, followed by further training on samples from this dataset with our methods applied. Additionally, we investigate one more scenario where we directly employ our method to reweight samples during the instruction tuning process, outside of a continual learning context.

 

\textbf{Training Datasets.} When conducting experiments on continual pre-training, we leverage the \texttt{C4}~\citep{raffel2020exploring} dataset as the source of training samples, aligning with the pre-training protocols of most LLMs. For experiments related to instruction tuning, we employ two widely-used datasets: (1) \texttt{Alpaca}~\citep{alpaca}, which contains 52K samples covering general tasks; and (2) \texttt{Open-Platypus}~\citep{lee2023platypus}, which contains data samples from more specific domains that are designed to improve models' performance on reasoning tasks. 

\textbf{Evaluation Datasets.} Our evaluation benchmarks cover multiple facets: (1) Commonsense reasoning: ARC challenge (ARC-C)~\citep{clark2018think}, HellaSwag~\citep{zellers2019hellaswag}, PiQA~\citep{bisk2020piqa}, and WinoGrande~\citep{sakaguchi2021winogrande}; (2) Reading comprehension: BoolQ~\citep{clark2019boolq}; and (3) Popular aggregated benchmarks: MMLU (including zero-shot - denoted as MMLU$^*$; and 5-shot)~\citep{hendrycks2020measuring}. We also report the average of scores calculated on these benchmarks.

\textbf{Training Settings.} For continual pretraining, we train each model using $100$ batches with a fixed learning rate of $5\times10^{-7}$ for \texttt{OPT} and $2\times 10^{-6}$ for \texttt{Sheared-LLaMA} and \texttt{LLaMA-v2}. For continual instruction-tuning, we similarly train models using $100$ batches with an initial rate of $3\times10^{-5}$. The optimizer we use for the experiments is AdamW~\citep{loshchilov2017decoupled}, with a weight decay of $0.01$. $r$ is set to $10$ for continual pretraining and instruction tuning experiments. We have provided an ablation study in Section~\ref{sec:ablation} to understand the effect of hyper-parameters.

\vspace{-0.2em}
\subsection{Experimental Results.}
\vspace{-0.2em}
\textbf{Continual Pretraining Comparison.} We first compare the performance of IR-DRO with baselines when continually pre-training the models. Table~\ref{tab:performance_on_opt_ir_dro} displays the results in multiple architectures, encompassing a wide spectrum of parameter counts.  The findings suggest that uniform weighting during the pretraining phase generally enhances performance, and MidRanking can further amplify this effect by up to 0.2 in terms of the average scores. Furthermore, IR-DRO consistently outperforms these baselines, achieving the best performance with an improvement of up to $0.60$ in average scores. Remarkably, such an improvement is achieved with just \textbf{100 more batches}, highlighting a modest computational effort, demonstrating promising room for enhancing the capabilities of existing pre-trained LLMs.


\textbf{Continual Instruction-Training Comparison.}
We continue to evaluate and compare IR-DRO and within the framework of continuous instruction tuning. Acquiring high-quality data for instruction tuning is highly valuable but costly, often requiring manual curation or solicitation from large-parameter models (\textit{e.g.}, GPT-4). This highlights the need to further utilize and leverage these datasets, as their creation requires significant effort and they can consequently be considered to be in short supply.
The models are first trained on \texttt{Open-Platypus} for $3$ epochs. Subsequently, these models undergo additional fine-tuning with re-weighted losses on randomly selected batches from the instruction-tuning dataset, in line with the aforementioned continual pre-training configurations.

Table~\ref{tab:performance_on_instruction} underscores the potential for improving model performance through sustained training on a consistent data set. A closer look at the table reveals a modest increase of $0.04$ in the average score for the MidRanking baseline, which serves to illustrate the constraints of manual hyperparameter optimization. In stark contrast, the implementation of IR-DRO leads to a significant uplift in model efficacy, evidenced by an upsurge of $0.46$ points in the average score, thereby outperforming all comparative baseline models. This improvement is particularly pronounced in the MMLU benchmarks, where IR-DRO registers impressive gains of $1.72$ and $0.55$ in zero-shot and 5-shot configurations, respectively.

\begin{table*}[ht]
    \centering
    \caption{Performance of \texttt{LLaMA-v2-7b} on multiple benchmarks under the setting of continual instruction-tuning: the models are first trained with a standard instruction tuning pipeline on the \texttt{Open-Platypus} dataset, and being continually fine-tuned with $100$ batches on the same dataset. ``Original'' indicates the experiment results with no continual instruction-tuning. }
     \label{tab:performance_on_instruction}
    \resizebox{0.95\linewidth}{!}{
    \begin{tabular}{c|c|ccccccc|c}
    \toprule \multirow{2}{*}{Training Set} & \multirow{2}{*}{Method} & \multicolumn{7}{c|}{Evaluation Benchmarks} & \multirow{2}{*}{Average} \\
     & & ARC-C & HellaSwag & PiQA & WinoGrande & BoolQ & MMLU & MMLU$^*$\\
        \midrule
        \multirow{5}{*}{\texttt{Open-Platypus}} 
        & Original & 47.18 & 76.30 & 78.78 & 68.75 & 78.78 &  48.20 & 44.56 & 63.22 \\ 
        & LowRanking & 47.48 & 76.40 & 78.69 & 68.39 & 79.19 & 49.00 & 45.86 & 63.57 \\
        & HighRanking & 45.48 & 76.30 & 78.78 & 68.75 & 79.05 & 47.69 & 44.41 & 62.92 \\
        & MidRanking & 46.17 & 76.25 & 78.81 & 68.66 & 79.13 & 48.13 & 45.70 & 63.26 \\
        & IR-DRO & 47.61 & 76.45 & 78.67  & 68.51 & 79.48 & 48.75 & 46.28 & \textbf{63.68} \\        
        \bottomrule
    \end{tabular}}
    \vspace{-1em}
\end{table*}

\textbf{Extension to Standard Instruction Tuning Experiments.}
Finally, we assess the efficacy of our proposed method for re-weighting samples during normal instruction tuning. This is to assess the flexibility of our algorithm beyond continual pre-training and continual instruction-tuning. In this case, the ``Uniform'' baseline corresponds to the widely accepted direct fine-tuning algorithm for instruction tuning.

\begin{figure}[ht]
    \centering
    \vspace{-1.5em}
    \includegraphics[width=1\linewidth]{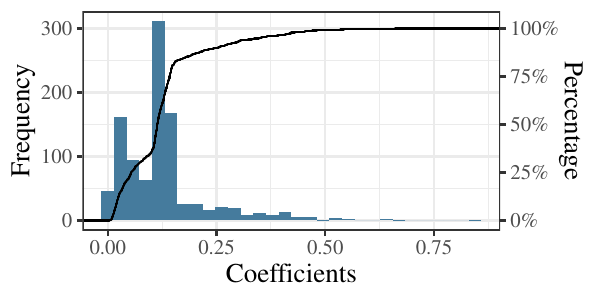}
    \vspace{-2.5em}
    \caption{The distribution of the coefficients generated by IR-DRO during continual pre-training. We visualize both the frequency of the coefficients and also the cumulative percentage. }
    \label{fig:coef_distribution}
    \vspace{-1em}
\end{figure}

We begin with a pre-trained \texttt{LLaMA-v2-7B} model and apply the algorithms to reweight or select samples thoroughly during the instruction tuning process. The results are detailed in Table~\ref{tab:performance_on_instruction_only}, revealing that: (1) selecting samples based solely on the highest or lowest losses can be too drastic, leading to a marked reduction in performance, especially in the MMLU benchmarks; (2) compared to the conventional ``Uniform'' approach, IR-DRO achieves better results, proving the effectiveness of our approach in improving performance after training on instruction-tuning datasets; (3) the impact of various methods varies across different benchmarks. For example, WinoGrande benefits more from training on simpler samples, whereas the MMLU benchmarks show a significant drop in performance when trained exclusively on such samples.

\begin{table*}[ht]
    \centering
    \caption{Performance of \texttt{LLaMA-v2-7b} on multiple benchmarks under the standard instruction tuning setting. The models are trained on \texttt{Alpaca} and \texttt{Open-Platypus} with our methods enabled, respectively. }
    \resizebox{0.95\linewidth}{!}{
    \begin{tabular}{c|c|ccccccc|c}
    \toprule \multirow{2}{*}{Training Set} & \multirow{2}{*}{Method} & \multicolumn{7}{c|}{Evaluation Benchmarks} & \multirow{2}{*}{Average} \\
    
     & & ARC-C & HellaSwag & PiQA & WinoGrande & BoolQ & MMLU & MMLU$^*$\\
    \midrule
        \multirow{5}{*}{\texttt{Alpaca}} 
        & Uniform & 47.78 & 75.58 & 79.71 & 69.61 & 78.76 &  47.41 & 43.58 & 63.20 \\ 
        & LowRanking & 46.67 & 75.58 & 79.98 & 69.46 & 78.62 &  46.90 & 42.29 & 62.79 \\
        & HighRanking & 47.70 & 75.52 & 79.60 & 69.53 & 78.99 & 46.24 & 42.20 & 62.83 \\
        & MidRanking & 47.72 & 75.54 & 79.71 & 69.59 & 78.92 & 46.61 & 42.31 & 62.91 \\
        & IR-DRO & 47.87 & 75.56 & 79.60 & 69.77 & 79.11 & 47.49 & 43.63 & \textbf{63.29}\\
        \midrule
        \multirow{5}{*}{\texttt{Open-Platypus}} 
        & Uniform & 47.18 & 76.30 & 78.78 & 68.75 & 78.78 &  48.20 & 44.56 & 63.22\\ 
        & LowRanking & 47.27 & 76.39 & 78.84 & 69.53 & 79.39 & 40.32 & 32.03 & 60.54 \\
        & HighRanking & 47.10 & 76.37 & 78.94 & 68.67 & 78.23 & 44.74 & 34.01 & 61.15 \\
        & MidRanking & 47.13 & 76.32 & 78.82 & 68.66 & 78.45 & 44.61 & 35.12 & 61.30 \\
        & IR-DRO & 47.10 & 76.32 & 78.84 & 68.72 & 78.87 & 48.60 & 45.08 & \textbf{63.36}\\        
        \bottomrule
    \end{tabular}}
    
    \label{tab:performance_on_instruction_only}
\end{table*}
\vspace{-0.5em}
\subsection{Ablations and Visualizations}
\vspace{-0.5em}

\label{sec:ablation}

\textbf{Distribution of Coefficients.} 
Figure~\ref{fig:coef_distribution} illustrates the distribution of coefficients and the cumulative percentage generated by IR-DRO across training batches. The coefficients exhibit significant variability, indicating that IR-DRO effectively prevents degradation into uniform assignment.

\textbf{Performance with different training settings.} 
In this section, we conduct ablation studies on hyperparameters used in continual pre-training, due to their significant impact on performance. We mainly examine the effects of the number of training steps (i.e., batches) and the learning rates.

\begin{figure}[ht]
    \centering
    \vspace{-0.5em}
    \includegraphics[width=0.9\linewidth]{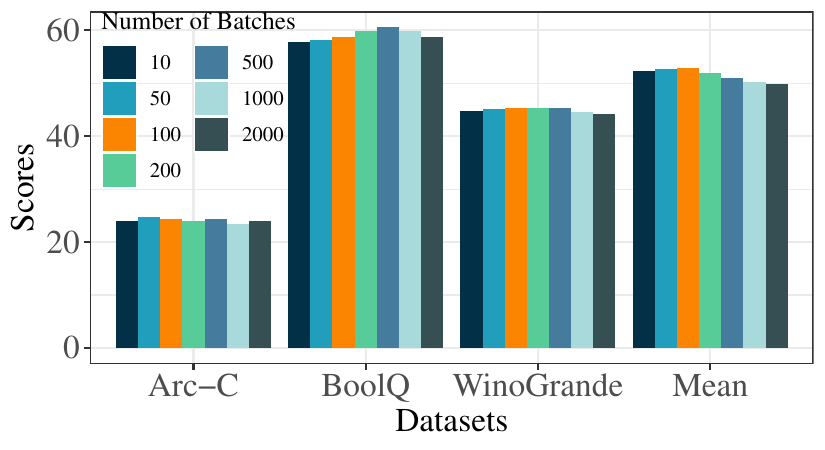}
    \vspace{-0.7em}
    \caption{Models' performance after training with different number of batches. We evaluate the continually trained \texttt{OPT-350M} using IR-DRO on three datasets (Arc-Challenge, BoolQ and WinoGrande), and report their average scores. }
    \vspace{-1.5em}
    \label{fig:ablation1}
\end{figure}

\begin{figure}[h!]
    \centering
    \vspace{-0.5em}
    \includegraphics[width=0.9\linewidth]{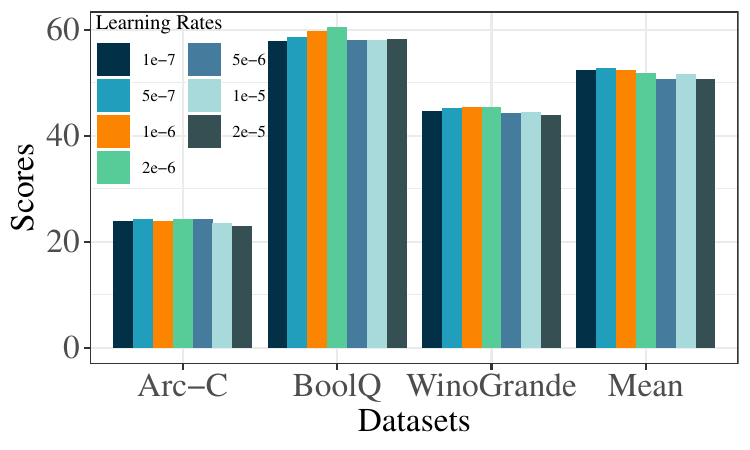}
    \vspace{-0.5em}
    \caption{Models' performance after training with different learning rates. We evaluate continually trained \texttt{OPT-350M} model using IR-DRO on three datasets (Arc-Challenge, BoolQ and WinoGrande), and calculate their average scores.}
    \vspace{-1.5em}
    \label{fig:ablation2}

\end{figure}

Our results are graphically represented in Figure~\ref{fig:ablation1} and \ref{fig:ablation2}, which depict the effects of our hyperparameter choices on model performance. These figures confirm that our hyperparameter settings are currently delivering satisfactory results. Additionally, they highlight that various evaluation benchmarks necessitate different extents of continual training. For instance, the BoolQ dataset reaches optimal performance after 500 training batches, in contrast to the WinoGrande dataset, which achieves its best results with only 100 training batches. This suggests the presence of an ideal juncture in the continual training trajectory, beyond which some benchmarks may not exhibit further performance gains. Figure~\ref{fig:ablation2} also shows a related trend regarding learning rates, where a higher rate appears to favor the BoolQ dataset, while other benchmarks tend to respond better to lower rates.

\vspace{-0.5em}
\section{Conclusion}
\vspace{-0.5em}

In this study, we introduce multiple strategies to pinpoint and utilize key samples from existing training datasets to augment the continuous training of LLMs. Beginning with an empirical approach, we devise an algorithm that selects samples with moderately high loss values, which are considered informative for model refinement. This is in contrast to the highest-loss samples that are typically excluded due to their association with noise and complexity. We then evolve this empirical strategy into a more structured framework based on distributionally robust optimization, which employs an instance-reweighting mechanism to dynamically focus training on informative samples. This framework is designed for seamless integration with current training protocols, featuring a closed-form solution that facilitates its application. Our comprehensive experiments across a variety of models and datasets demonstrate that our targeted sampling methods significantly boost LLM performance on multiple benchmarks, at lightweight costs.




\bibliography{example_paper}
\bibliographystyle{icml2024}

\newpage
\appendix
\onecolumn

\section{Proof of the optimal weights expression} \label{app:proof}
In order to solve the maximization problem, we
will fix $\boldsymbol{\theta}$ and derive an optimal solution $\boldsymbol w^*(\boldsymbol{\theta})$ that depends on $\boldsymbol{\theta}$. To this end, we reformulate the `max' problem as follows: 
\begin{align*}
    \min_{\boldsymbol{w} \in \mathcal{P}} - \sum_{i=1}^N w_i \mathcal{L}(\boldsymbol{\theta}; \boldsymbol{x}_i) + r \sum_{i=1}^N w_i \log(Nw_i)
\end{align*}
Due to the fact that $\sum_{i=1}^N w_i = 1$, we have $r \sum_{i=1}^N w_i \log(Nw_i) = r \sum_{i=1}^N w_i \log(w_i) + r \log(N)$. Thus, the problem becomes
\begin{align*}
    \min_{\boldsymbol{w} \in \mathcal{P}} - \sum_{i=1}^N w_i \mathcal{L}(\boldsymbol{\theta}; \boldsymbol{x}_i) + r \sum_{i=1}^N w_i \log(w_i), 
\end{align*}
where we drop the term $r \log(N)$, which is independent of $\boldsymbol{w}$. 
There are three constraints to handle, i.e., $w_i \geq 0$; and $w_i \leq 1$; $\sum_{i=1}^N w_i = 1$.
Note that the constraint $w_i \geq 0$ is enforced
by the term $w_i \log(w_i)$, otherwise the above objective will become infinity. As a result, the constraint
$w_i < 1$ is automatically satisfied due to $\sum_{i=1}^N w_i = 1$ and $w_i \geq 0$. Hence, we only need to explicitly
tackle the equality constraint $\sum_{i=1}^N w_i = 1$. To this end, we define the following Lagrangian function. 
\begin{align*}
    L_{\boldsymbol{\theta}} (\boldsymbol{w}, \lambda) = - \sum_{i=1}^N w_i \mathcal{L}(\boldsymbol{\theta}; \boldsymbol{x}_i) + r \sum_{i=1}^N w_i \log(w_i) + \lambda \left(\sum_{i=1}^N w_i - 1\right), 
\end{align*}
where $\lambda$ is the Lagrangian multiplier for the equality constraint $\sum_{i=1}^N w_i = 1$. From the KKT conditions, the optimal solutions $\boldsymbol{w}^*$ must satisfy
\begin{align*}
    - \mathcal{L}(\boldsymbol{\theta}; \boldsymbol{x}_i) + r\left(\log(w_i^*\right) + 1) + \lambda = 0 \quad \text{ and } \quad \sum_{i=1}^N w_i^* = 1
\end{align*}
From the first equation, we can derive $w^*_i = \frac{1}{C} \exp\left(\frac{\boldsymbol {\mathcal{L}}_i(\boldsymbol{\theta})}{r}\right)$, where $C$ is a constant. Due to the second equation, we can conclude that $C$ is given by $\sum_j \exp\left(\frac{\boldsymbol {\mathcal{L}}_j(\boldsymbol{\theta})}{r}\right)$ and that $w^*_i = \frac{\exp\left(\frac{\boldsymbol {\mathcal{L}}_i(\boldsymbol{\theta})}{r}\right)}{\sum_j \exp\left(\frac{\boldsymbol {\mathcal{L}}_j(\boldsymbol{\theta})}{r}\right)}$, which completes the proof. 



\end{document}